%% file: main.tex
\documentclass[conference, compsoc]{IEEEtran}
\IEEEoverridecommandlockouts
\usepackage{epsfig, booktabs, color, soul, soulpos, multirow, adjustbox, pgfplots, balance,CJKutf8,amsfonts,amsmath}

\newenvironment{definition}[1][Definition]{\begin{trivlist}
\item[\hskip \labelsep {\bfseries #1}]}{\end{trivlist}}

\ulposdef{\hlcyan}[xoffset=1pt]{\mbox{\color{cyan!30}\rule[-.8ex]{\ulwidth}{3ex}}}
\ulposdef{\hlred}[xoffset=1pt]{\mbox{\color{red!30}\rule[-.8ex]{\ulwidth}{3ex}}}
\ulposdef{\hlyellow}[xoffset=1pt]{\mbox{\color{yellow!30}\rule[-.8ex]{\ulwidth}{3ex}}}
\ulposdef{\hlorange}[xoffset=1pt]{\mbox{\color{orange!30}\rule[-.8ex]{\ulwidth}{3ex}}}
\ulposdef{\hlgreen}[xoffset=1pt]{\mbox{\color{green!30}\rule[-.8ex]{\ulwidth}{3ex}}}

\begin{document}
% \title{Inferring Concept Transitions from Medical Text Queries via Energy-based RNN-CNN Neural Networks}
% \title{Inferencing Medical Concept Transitions from Online Healthcare Questions}
\title{Bringing Semantic Structures to User Intent Detection in Online Medical Queries}
\if{false}
\author{Chenwei Zhang}
\email{czhang99@uic.edu}
\affiliation{\institution{University of Illinois at Chicago}\city{Chicago}\state{IL}\country{USA}}
\author{Nan Du}
\email{nandu@baidu.com}
\affiliation{\institution{Baidu Research Big Data Lab}\city{Sunnyvale}\state{CA}\country{USA}}
\author{Wei Fan}
\email{fanwei03@baidu.com}
\affiliation{\institution{Baidu Research Big Data Lab}\city{Sunnyvale}\state{CA}\country{USA}}
\author{Yaliang Li}
\email{yaliangli@baidu.com}
\affiliation{\institution{Baidu Research Big Data Lab}\city{Sunnyvale}\state{CA}\country{USA}}
\author{Chun-ta Lu}
\email{clu29@uic.edu}
\affiliation{\institution{University of Illinois at Chicago}\city{Chicago}\state{IL}\country{USA}}
\author{Philip S. Yu}
\affiliation{\institution{University of Illinois at Chicago}\city{Chicago}\state{IL}\country{USA}}
\affiliation{\institution{Tsinghua University}\city{Beijing}\country{China}}
\email{psyu@uic.edu}  
\fi

\author{\IEEEauthorblockN{Chenwei Zhang\IEEEauthorrefmark{1}\IEEEauthorrefmark{5}, Nan Du\IEEEauthorrefmark{2}, Wei Fan\IEEEauthorrefmark{3}\IEEEauthorrefmark{6}, Yaliang Li\IEEEauthorrefmark{2}, Chun-Ta Lu\IEEEauthorrefmark{1}, Philip S. Yu\IEEEauthorrefmark{1}\IEEEauthorrefmark{4}}
\thanks{\IEEEauthorrefmark{5} Part of the work was done when the author was an intern at Baidu Research Big Data Lab.}
\thanks{\IEEEauthorrefmark{6} Part of the work was done when the author was employed by Baidu Research Big Data Lab.}

\IEEEauthorblockA{\IEEEauthorrefmark{1}Department of Computer Science, University of Illinois at Chicago, Chicago, IL, USA\\
\IEEEauthorrefmark{2}Baidu Research Big Data Lab, Sunnyvale, CA, USA\\
\IEEEauthorrefmark{3}Tencent Medical AI Lab, Palo Alto, CA, USA\\
\IEEEauthorrefmark{4}Institute for Data Science, Tsinghua University, Beijing, China\\
Email: \IEEEauthorrefmark{1}\{czhang99,clu29,psyu\}@uic.edu, \IEEEauthorrefmark{3}davidwfan@tencent.com, \IEEEauthorrefmark{2}\{nandu, yaliangli\}@baidu.com}
}

% \author{
%  Chenwei Zhang$^\dag$
%  \mbox{\hspace{.1in}}
%  Nan Du$^\natural$
%  \mbox{\hspace{.1in}}
%  Wei Fan$^\natural$
%  \mbox{\hspace{.1in}}
%  Yaliang Li$^\natural$
%  \mbox{\hspace{.1in}}
%  Chun-ta Lu$^\dag$
%  \mbox{\hspace{.1in}}
%  Philip S. Yu$^{\dag\S}$}
% \affiliation{$^\dag$Department of Computer Science, University of Illinois at Chicago, Chicago IL, USA}
% \affiliation{\vspace{-0.15in}$^\natural$Baidu Research Big Data Lab, Sunnyvale CA, USA}
% \affiliation{$^\S$Institute for Data Science, Tsinghua University, Beijing, China}
%  \affiliation{\vspace{-0.1in}$^\dag$\{czhang99,clu29,psyu\}@uic.edu, $^\natural$\{fanwei03,nandu, yaliangli\}@baidu.com}

\maketitle
\begin{abstract}
The Internet has revolutionized healthcare by offering medical information ubiquitously to patients via web search. 
% With the ever-increasing in dire need of healthcare professionals around the world, more and more patients with inconvenient accesses to offline medical resources turn to the world wide web for quick and reliable answers.
% The online healthcare services changed the way people obtain professional medical 
% users ubiquitous access to medical information.
% Online medical services offer abundant medical knowledge to users. 
% With an unprecedented blooming of healthcare web search, 
% Medical-related questions that are posted and extensively searched online are intended for specific healthcare information needs.
% Among the massive online medical text questions, 
% Concept transitions indicating how users' information needs are expressed and developed by mentioning different medical concepts. 
The healthcare status, complex medical information needs of patients are expressed diversely and implicitly in their medical text queries.
Aiming to 
% bridge the diversely expressed medical text queries with structured semantic transitions, 
better capture a focused picture of user's medical-related information search
and shed insights on their healthcare information access strategies, it is challenging yet rewarding to detect structured user intentions from their diversely expressed medical text queries.
% To meet this goal, we observe and study user concept transitions in real-world medical text queries.
We introduce a graph-based formulation to explore structured concept transitions for effective user intent detection in medical queries, where each node represents a medical concept mention and each directed edge indicates a medical concept transition.
% An end-to-end deep model is proposed to extract structured semantic transitions from each medical text query. 
% More specifically, an Given a diversely expressed real-world medical text query, 
% concept transitions are modeled on a directed concept graph where each node represents a concept mention and each directed edge indicates a medical concept transition. 
% A novel neural network model is introduced in this work, which gives an end-to-end solution to the concept transition inference problem. 
A deep model based on multi-task learning is introduced to extract structured semantic transitions from user queries, where the model extracts word-level medical concept mentions as well as sentence-level concept transitions collectively. 
% an in to a strutred model detects medical concept mentions, and is able to infer a full spectrum of medical concept transitions among those concept mentions on real-world healthcare questions collectively. 
A customized graph-based mutual transfer loss function is designed to impose explicit constraints and further exploit the contribution of mentioning a medical concept word to the implication of a semantic transition. We observe an 8\% relative improvement in AUC and 23\% relative reduction in coverage error by comparing the proposed model with the best baseline model for the concept transition inference task on real-world medical text queries.
% The source code of this work is available online\footnote{\href{https://github.com/czhang99/CTI}{https://github.com/czhang99/CTI}}.
\end{abstract}

% \renewcommand{\shortauthors}{C. Zhang et al.}
%\keywords{Concept Transition; Medical Query; Information Search; Neural Network; Data Mining}

\input{subfiles/1_introduction.tex}

\input{subfiles/2_problem_formulation.tex}

\input{subfiles/3_proposed_model.tex}

\input{subfiles/4_experiments.tex}

\input{subfiles/5_related_works.tex}

\input{subfiles/conclusion.tex}
%\end{document}  % This is where a 'short' article might terminate

%ACKNOWLEDGMENTS are optional
% \section{Acknowledgments}
% This section is optional; it is a location for you
% to acknowledge grants, funding, editing assistance and
% what have you.  In the present case, for example, the
% authors would like to thank Gerald Murray of ACM for
% his help in codifying this \textit{Author's Guide}
% and the \textbf{.cls} and \textbf{.tex} files that it describes.

% \section{Acknowledgments}
% This work is supported in part by NSF through grants IIS-1526499, and CNS-1626432, and NSFC 61672313.


\bibliographystyle{plain}
\balance\bibliography{sigproc}  % sigproc.bib is the name of the Bibliography in this case

% ACM needs 'a single self-contained file'!
%\balancecolumns

\end{document}

%% file: subfiles/1_introduction.tex
\section{Introduction}
The shortages of healthcare professionals are leading to healthcare systems plagued by bottlenecks. According to the World Health Organization, the world will face a shortfall of nearly 13 million healthcare professionals by 2035 \cite{campbell2013universal}. 
% Conventional offline delivery of medical services is greatly limited by schedules of healthcare professions, locations, and other physical requirements. What's more, due to an unbalanced geographic distribution of medical human resources, an increasing number of patients with medical-related information needs may not have convenient and reliable access to offline, physical medical resources.
In the meanwhile, an increasing number of medical-related online services emerge on the world wide web to offer ubiquitous medical information to patients via their web search \cite{fox2013one}. For example, the Chinese search engine Baidu processes over 6 billion search queries every day, while 60 million of them are healthcare-related text queries\footnote{http://science.china.com.cn/2016-11\/24\/content\_9180719.htm}. Online medical question answering forums such as xywy.com\footnote{http://club.xywy.com} has 120 million registered users and more than 22 million unique daily visitors. 

With the flourishing demand for medical-related services, it is crucial for service providers to infer implicit user intentions from the diversely expressed medical text question: what medical concepts a user mentions and how concept transitions are formulated among these concepts.
% With the booming development of the world wide web, the internet has more fundamental impacts on alleviating the shortage of medical professionals around the world. The world wide web enables a universal accessibility of high-quality online medical-related services globally. Many online medical question-answering websites or medical knowledge search engines emerges for such purpose, on which users (usually patients) post medical-related questions and healthcare professionals can give advice and suggestions directly online responsively. The accessibility and availability of online medical services via the world wide web dramatically reinforce the universal health coverage.
% What is medical query?\\
Generally, medical text queries that users search online or post on medical question-answering websites express various medical-related conditions and indicate different information needs, as shown in Table \ref{tab::medical_queries}.

\begin{table}[bth!]
    \centering
    \begin{tabular}{p{3.in}}
        $\circ$ Medical Text Questions\\
        $\bullet$ Inferred Concept Mentions \& Concept Transitions\\
        \toprule
        $\circ$ \hlred{Why} do I get \hlcyan{dizzy} so often?\\
        $\bullet$ \textit{\hlcyan{Symptom}} $\to$ \textit{\hlred{Cause}}\\ \hline
        $\circ$ My three-year-old child is \hlcyan{sick with a temperature of 100 degrees} she \hlcyan{can't keep anything down including liquids}. \hlyellow{What kind of medicine} should I give my child, \hlgreen{and how much}? \\
        $\bullet$ \hlcyan{Symptom} $\to$ \textit{\hlyellow{Medicine}} $\to$ \textit{\hlgreen{Instruction}}\\ \hline
        $\circ$ Do I have \hlorange{insomnia} if I have \hlcyan{trouble staying asleep}? Any \hlyellow{medication} is recommended to help me \hlcyan{fall asleep } easier?\\
        $\bullet$ \textit{\hlorange{Disease}} $\leftarrow$ \textit{\hlcyan{Symptom}} $\to$ \textit{\hlyellow{Medicine}}\\
        \bottomrule
    \end{tabular}
     \caption{Medical queries and the extracted medical concept mentions \&transitions.}
%     Phrases in text queries  are marked by different colors for illustration purposes.}
    \label{tab::medical_queries}
\vspace{-2mm}
\end{table}

Usually, medical semantic transitions are formulated by users during their efforts to express their existing medical conditions as well as their intended medical-related information needs, either explicitly or implicitly. The diversely expressions cover the mention of different types of medical concepts, each represents a set of notions such as symptoms, diseases, medicines etc. In real-world medical text queries, various expressions can be referred to as a concept mention, either explicitly (e.g. ``Tylenol'', ``Ibuprofen'' for the \textit{medicine} concept) or implicitly (such as ``what'', ``which drug/medicine''). Even for the same medical concept, different expressions can be used. For example, ``nose plugged'', ``blocked nose'' and ``sinus congestion'' all belong to the same symptom concept mention but expressed very differently.
% A lot of recent works try to learn word-concept mapping from queries.

% What is concept transition\\
% {\color{red}mention where the original transition graph comes from?}
The way concept mentions organized in a question naturally forms a structured concept transition graph that reflects users information-seeking intentions. Such ubiquitous observation in medical text questions is rarely studied in previous literatures. A typical formulation for medical intent detection is to model each semantic transition as a single label \cite{zhang2014classification}, or as a two-element tuple indicating 1) what a user have described and 2) what information the user is looking for \cite{zhang2016mining}. For example, we can have $(Symptom, Disease)$ and $(Symptom, Medicine)$ for the last query in Table \ref{tab::medical_queries}. This formulation defines each tuple as an individual label and ignores the correlations among the multiple semantic transitions in a single query. In real-world medical text queries, multiple semantic transitions in a single question may conjugate with each other by mentioning the same medical concept. For example, $(Symptom, Disease)$ and $(Symptom, Medicine)$ share the same concept $Symptom$ by expressing symptoms: ``trouble staying asleep'' and ``fall asleep'' in the query. 
% space
The above formulations fail to consider the semantic interactions among multiple medical concepts in a medical query, which prevent them from satisfactorily detect sophisticated user intentions with complex semantic structures.

Alternatively, we can formulate concept transitions over a directed, highly structured concept graph where concept mentions are nodes and transitions between concepts are directed edges between them. For example, 
% the second query in Table \ref{tab::medical_queries} mentions concepts ``Symptom'', ``Medicine'' and ``Instruction''. 
with a graph-based formulation, the concept transition for the second question in Table \ref{tab::medical_queries} is formulated as $Symptom \to Medicine \to Instruction$ since the user first describes his/her symptoms (``sick'', ``temperature of 100 degrees'') and inquires about information on medicine concepts (``What kind of medicine''), followed by phrases (``and how much'') indicating further information seek intentions about instructions on the medicine. Real-world text questions often exhibit a mixture of multiple concept transitions in each question (See Section \ref{sec::ob} for details), in which shared concept mentions serve as a bridge coupling two or more concept mentions into a structured concept transition. Thus, a graph-based formulation would essentially allow us to jointly model and infer correlations between concept mentions and multiple concept transitions simultaneously, which is one of our key contributions. 

\begin{figure}[hb]
\centering
\epsfig{file=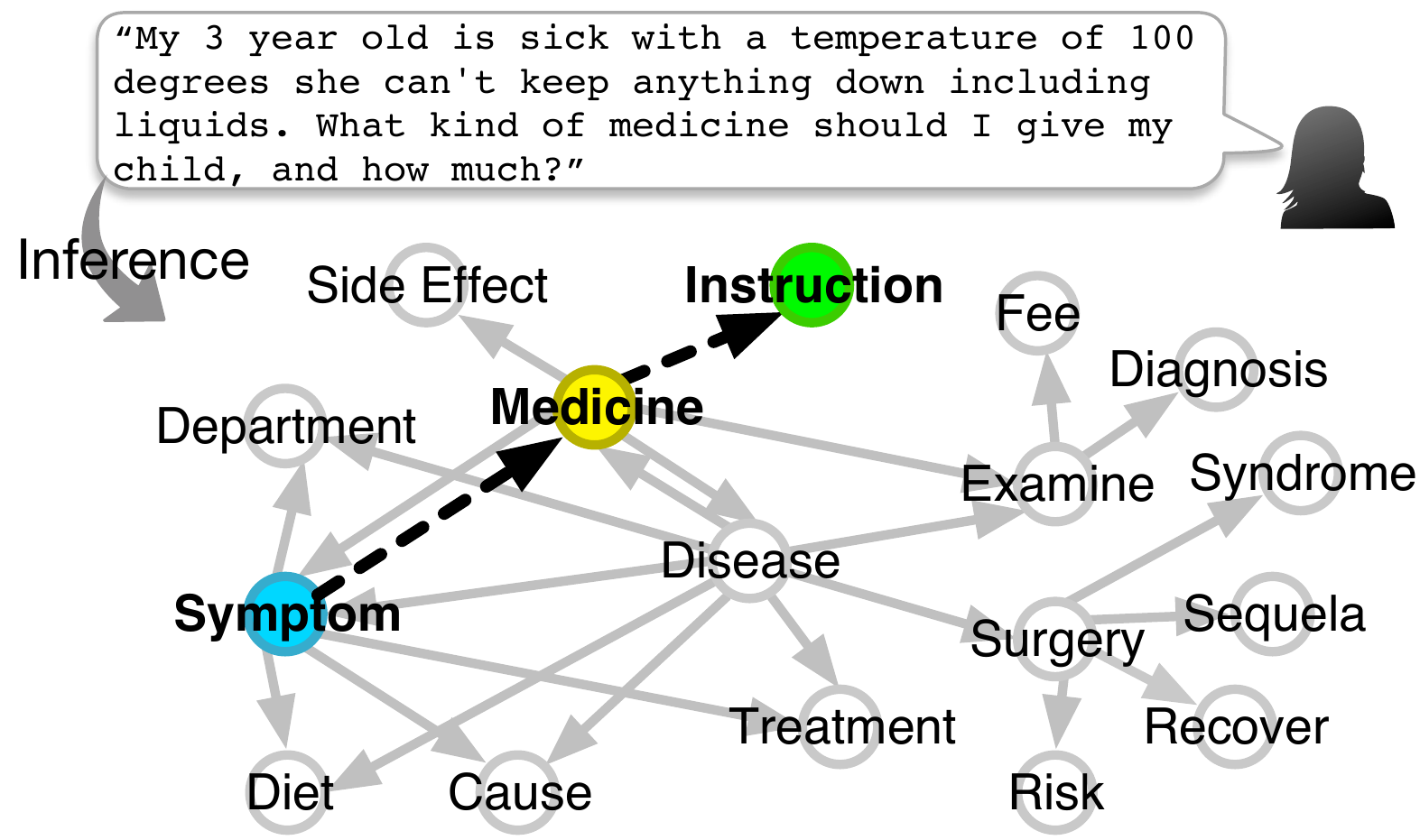, width=3.33in}
\caption{The concept transition inference problem over the full concept graph. Each node in the concept graph represents a concept. Each directed edge between two nodes indicates an information-seeking transition over two concepts. Given a text question and an existing medical concept transition graph, the concept transition inference problem extracts a directed, structured subgraph consisting of a set of concepts being mentioned (shown as colored nodes) as well as a set of concept transitions being encoded (shown as black dashed lines) from the question.}\label{fig::concept_transition_inference_problem}
\end{figure}

\noindent\textbf{Problem Studied}: In order to better capture a focused picture of people's medical-related information search and information access strategies, we propose and study the concept transition inference problem for online healthcare questions with a graph-based formulation.
% In this work, we obtained a concept transition graph from medical experts, as shown in Figure \ref{fig::concept_transition_inference_problem}. 
Given a question and a concept graph indicating the full spectrum of concept transitions in the medical domain, our goal is to effectively infer concept transitions that are activated by the given medical text question, as shown in Figure \ref{fig::concept_transition_inference_problem}.

% What is the challenge\\
\noindent\textbf{Challenges}:
A typical solution for the concept transition inference problem evolves hand-engineering features based on expert knowledge in the medical domain, such as constructing a word-concept mapping dictionary \cite{zhang2016mining} or using pre-defined rules \cite{de2010rule} or templates \cite{spink2004study} for question intent classification . Even if one discounts the tedious effort required for feature engineering, those features are usually designed for a limited number of questions accessible to domain experts and do not generalize to handle various user expressions in real-world medical text questions.  People with different knowledge background tends to express the same idea in different ways. For example, a medicine concept can be mentioned by specific drug names such as ``Tylenol'', ``Ibuprofen'' or phrases like ``what kind of medicine/drug/medication''. The decent performance of those approaches usually comes at the cost of acquiring an external knowledge base to handle varying linguistic modalities and diversified expressions. How to minimize feature engineering without compromising the performance for the concept transition inference is still challenging.

Moreover, comparing with general-purpose text questions which people have been posting or searching for online, where users only focus on a single concept (such as ``weather'', ``politics'' or ``stocks''), concept transitions in medical questions usually involve multiple concepts. It would take strenuous efforts to model correlations among multiple concept transitions without considering the shared concept mentions effectively. What's more, unlike many existing works on medical text analysis such as sentiment classification \cite{greaves2013use,ali2013can} which consider positive, negative or neutral sentiments in medical texts, it is challenging yet rewarding to consider structured concept transitions which model sophisticated medical semantic transitions in real-world medical text queries.

\noindent\textbf{Proposed Work:} 
To overcome those challenges, we introduce a novel neural network architecture that bring structures to semantic transitions for user intent detection in medical text queries. We observe an appealing property that real-world medical text questions exhibit a strong coupling between concept mentions and concept transitions. Consequently, a graph-based formulation is defined to jointly model correlations between concept mentions and multiple concept transitions. The proposed model can effectively infer concept transitions from real-world medical text queries and extract a structured representation of users information-seeking intentions. 

% space
% Once we figure out the structured concept transitions in a question, various traditional tasks in web search \cite{dai2011enhancing} such as query rewriting \cite{zhao2014query}, click prediction \cite{hu2011characterizing}, question-answering \cite{yang2015beyond} and recommendation \cite{zhai2016deepintent,duan2015mining} where intentions are previously represented as fine-grained labels may now benefit from this graph-based formulation. 
% Not to mention future tasks specifically target on graph-based concept transitions.

Also, we introduce an end-to-end solution to the concept transition inference problem which is well integrated with the graph-based concept transition formulation. The concept transition inference task is formulated
% as an optimization problem on a neural network structure which is designed to collectively infer concept mentions and concept transitions. The 
with a multi-task learning schema that learns to extract concept mentions as well as to infer concept transitions collectively. A customized graph-based mutual transfer loss function is designed to impose explicit constraints to reduce the conflicts between the concepts being mentioned and the concept transitions being activated. 

Furthermore, the proposed model minimizes hand-feature engineering by using only the text information from the medical text query and an existing medical concept graph introduced in \cite{zhang2016mining}, without relying on other external knowledge bases. The neural network is trained to automatically discover concept mentions and infer concept transitions from raw text questions, in contrast to relying on fixed dictionaries for word-concept mapping \cite{chiang2012autodict,godbole2010building,zhang2016mining} or using pre-defined parsing rules \cite{de2010rule} or templates \cite{spink2004study} in prior works. The neural network also learns to assign confidence scores to words as an attention mechanism \cite{zhai2016deepintent}, which makes the model self-explaining in indicating the contribution of each medical concept mention to the structured semantic transition.

Moreover, the proposed method learns both semantic and syntax representations for each word and its Part-of-Speech tag respectively. The learned embeddings are fed into two separate recurrent neural networks to build a memory summarizing multiple concept transitions over the input text sequence.
This compositionality of input embedding lends the proposed method to handle diversified expressions in user questions.

% Then, in order to deal with multiple concepts and transitions in a single question, a concept encoder is designed to encode multiple concepts mentioned in a variable-length sequence learned from RNNs to a sequence of concept vectors. While a transition encoder learns from the last hidden state of the RNN, which can be considered as a memory summarizing multiple concept transitions over the sequence. Finally, 

Experiments are conducted on real-world medical text queries collected from a medical question-answering forum, which is publicly available. We contrast the performance of the proposed model with other alternatives by an 8\% improvement in micro-AUC and an 23\% reduction in coverage loss for the concept transition inference problem.

Overall, our paper makes the following contributions:
\begin{enumerate}
\item We observe and formally define concept transitions in medical text questions and show appealing properties among concept transitions and shared concept mentions. 
\item We study the concept transition inference problem with a graph-based formulation, which brings semantic structures to diversely expressed natural language queries. 
\item We propose an end-to-end solution with a novel neural network model
% \footnote{Source code will be made available upon acceptance of the manuscript.} 
to the concept transition inference problem without excessive external knowledge requirements.
% : a novel neural network structure along with a medical concept graph without relying on domain-specific knowledge bases such as medical concept dictionaries.
\item We collect datasets and empirically evaluate the proposed method on real-world medical text queries.
\end{enumerate}

%% file: subfiles/2_problem_formulation.tex
\section{Preliminaries}
We now formally define the terminologies and describe the concept transition inference problem. Also, we provide observations to show appealing coupling properties of concept mentions and concept transitions in real-world text queries, which motivates a graph-based concept transition formulation.

\subsection{Terminologies}
\begin{definition}[Concept]
Let a concept $c$ be a group or class of objects and/or abstract ideas representing similar fundamental characteristics in a certain domain. $C=\{c_1,c_2,...,c_M\}$ is list of a full spectrum of $M$ concepts in a specific domain. ( e.g. the medical domain contains concepts of diseases, symptoms, medicine and so on). Users can mention concepts in a query by mentioning specific object names as explicit mentions (``Tylenol'', ``Ibuprofen'' or ``xxx caplet/capsule/drop/syrup''), as well as implicit mentions by abstract ideas that refer to concept (e.g. ``remedy'') or phrases indicating this concept (e.g. ``which medication/medicine/drug'').
\end{definition}

\begin{definition}[Concept Transition]
Let a concept transition ${t}_{{i}\to{j}}$ defines a transition of a user information search intent from the concept $c_i$ to concept $c_j$. A concept transition ${t}_{{i}\to{j}}$ exists in a query when two concepts $c_i$, $c_j\in{C}$ are mentioned (either explicitly or implicitly) with a semantic transition between them. For example, medical queries with concept transitions $t_{Symptom\to{Medicine}}$ usually start with patients describing their symptoms and asking for related information about medications that help them alleviate their symptoms.
% such as ``I got a temperature with 100 degrees, should I take the Tylenol?''. 

$T$ contains the full spectrum of $N$ concept transitions in a certain domain, which can be indexed as $T=\{t_1,t_2,...,t_N\}$ for simplicity instead of $\{t_{i\to{j}}\}$. Those two index notations are used interchangeably in this paper. Multiple concept transitions can be activated by a single query and the direction of a concept transition does not necessarily follow the order of concepts being mentioned in a query. Multiple concept transitions in a query may follow a natural chain-like path, such as the path $Symptom \to Medicine \to Instruction$.
\end{definition}

\begin{definition}[Concept Graph]
Let $G=\langle{C},{T}\rangle$ be a concept graph where each node represents a concept $c_m\in{C}$ and $t_{i,j}\in{T}$ be a directed edge from node $c_i$ to $c_j$. A concept graph $G$ is a graph representation of all possible concepts and concept transitions in a certain domain. Note that the domain-specific concept graph can be obtained from domain experts, which we adopted in this paper, or constructed from large text corpora by existing techniques \cite{hasegawa2004discovering,yan2009unsupervised,zhang2016heer}.
\end{definition}

\begin{definition}[Active Concept Graph]
Let an active concept graph $G_Q=\langle{C_Q, T_Q}\rangle$ be a subgraph of $G=\langle{C},{T}\rangle$, indicating concepts $ C_Q\subseteq C$ mentioned by a query $Q$ and concept transitions $T_Q \subseteq T$ activated by the the query $Q$.
\end{definition}

\subsection{Problem Statement}
% our goal is to effectively infer the concept transitions among those concepts over a concept transition graph.
% We now define the problem where given a text query in which one or multiple concepts are mentioned and concept transitions are expected to be inferred effectively.
\begin{definition}[The Concept Transition Inference Problem:]
Given 1) a text query $Q$ which consists of $K$ elements $\{q_1, q_2, ..., q_K\}$, where each element is a word or a phrase and 2) a concept graph $G=\langle C, T\rangle$, where $C$ denotes all possible concepts and $T$ indicates all possible concept transitions, the concept transition inference problem tries to effectively infer an active concept graph $\hat{G}_Q=\langle \hat{C}_Q,\hat{T}_Q\rangle$ given a query $Q$. Figure \ref{fig::concept_transition_inference_problem} illustrates this idea. 
\end{definition}

\subsection{Observations}\label{sec::ob}
Based on the terminologies and the problem defined, we would like to observe the existence of active concept graphs given real-world medical text queries. We sample 10,000 medical text queries from an online medical question answering forum and label them with concept transitions being activated. We end up having 17 unique types of concepts and 23 unique types of concept transitions (details in Section \ref{sec::data}).
\begin{table}[h]
\centering
\begin{adjustbox}{width=\columnwidth}
\begin{tabular}{llllll}
\toprule
Medical Concept & Symptom & Disease & Cause & Medicine & Treatment\\ \hline
Frequency & 7650  & 7446    & 5380  & 3733       & 2504\\
\bottomrule
\end{tabular}
\end{adjustbox}
\caption{Top frequent concepts mentions.}\label{tab::freq_concepts}
\end{table}
Table \ref{tab::freq_concepts} shows the top frequent concepts that are mentioned either explicitly or implicitly in medical text queries. 
% From the result we can see that symptom, disease, cause, medicine and treatment are the top five concepts that involves in concept transitions.

\begin{figure}[bth]
\centering
\epsfig{file=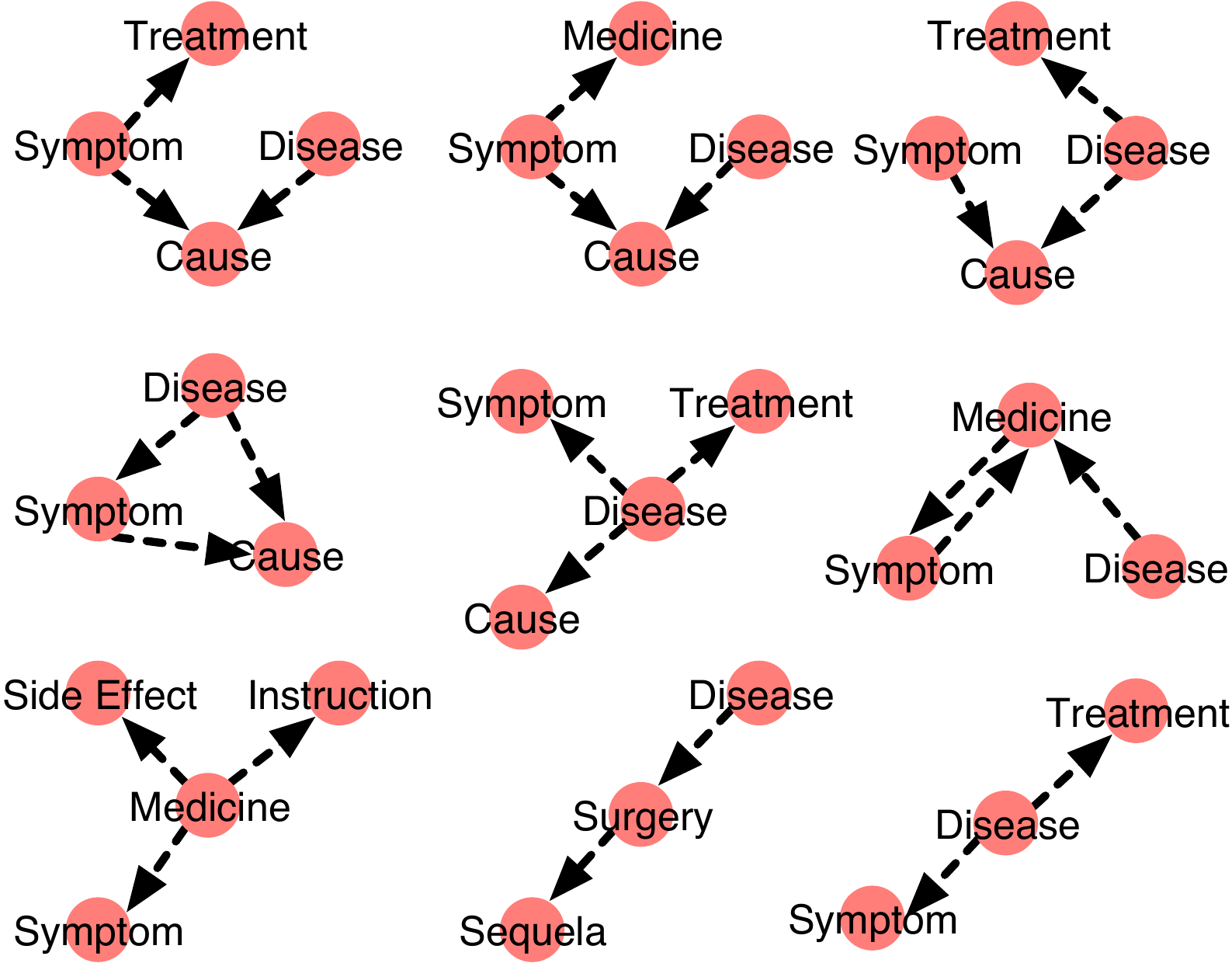, width=3.3in}
\caption{Popular active concept graphs.}\label{fig::top_concept_transitions}
\end{figure}

% Top 5 most frequent concepts transitions\\
% Show the graph where edge width is the frequency of concept transition\\
We also show 9 popular active concept graphs in  medical text queries, shown in Figure \ref{fig::top_concept_transitions}. By characterizing concept transitions with a graph-based formulation, it maps natural language queries with diversified expressions into a structured form, which show users information needs in a structured way.

More importantly, we found that active concept graphs rarely have disconnected components, from a perspective of the graph theory. This not only implies that users tend to have multiple concept transitions within a single medical text query but also indicates that multiple concept transitions in the same query are expressed and developed together, coupled with some shared concept mentions.
% In summary, characterizing the concept transition as a graph makes it easier...
The connectivity of active concept graphs implies that by taking advantages of the concepts and concept transitions formulated on a concept graph, we may able to utilize the correlations between nodes and edges for a better inference and therefore, users information search intent or information access strategy for their healthcare conditions can be modeled and inferred more effectively.
% For example, an active concept graph is more likely to be correct when it has

% From the perspective of link prediction problems on the relational data, the concept transition inference problem studied in this paper can be considered as an accurate determination of 1) a set of nodes representing concepts and 2) directed edges representing concept transitions from a concept transition graph that have been activated by a given query.

%% file: subfiles/3_proposed_model.tex
\section{Medical Concept Transition Inference}
\begin{figure}[t!]
\centering
\epsfig{file=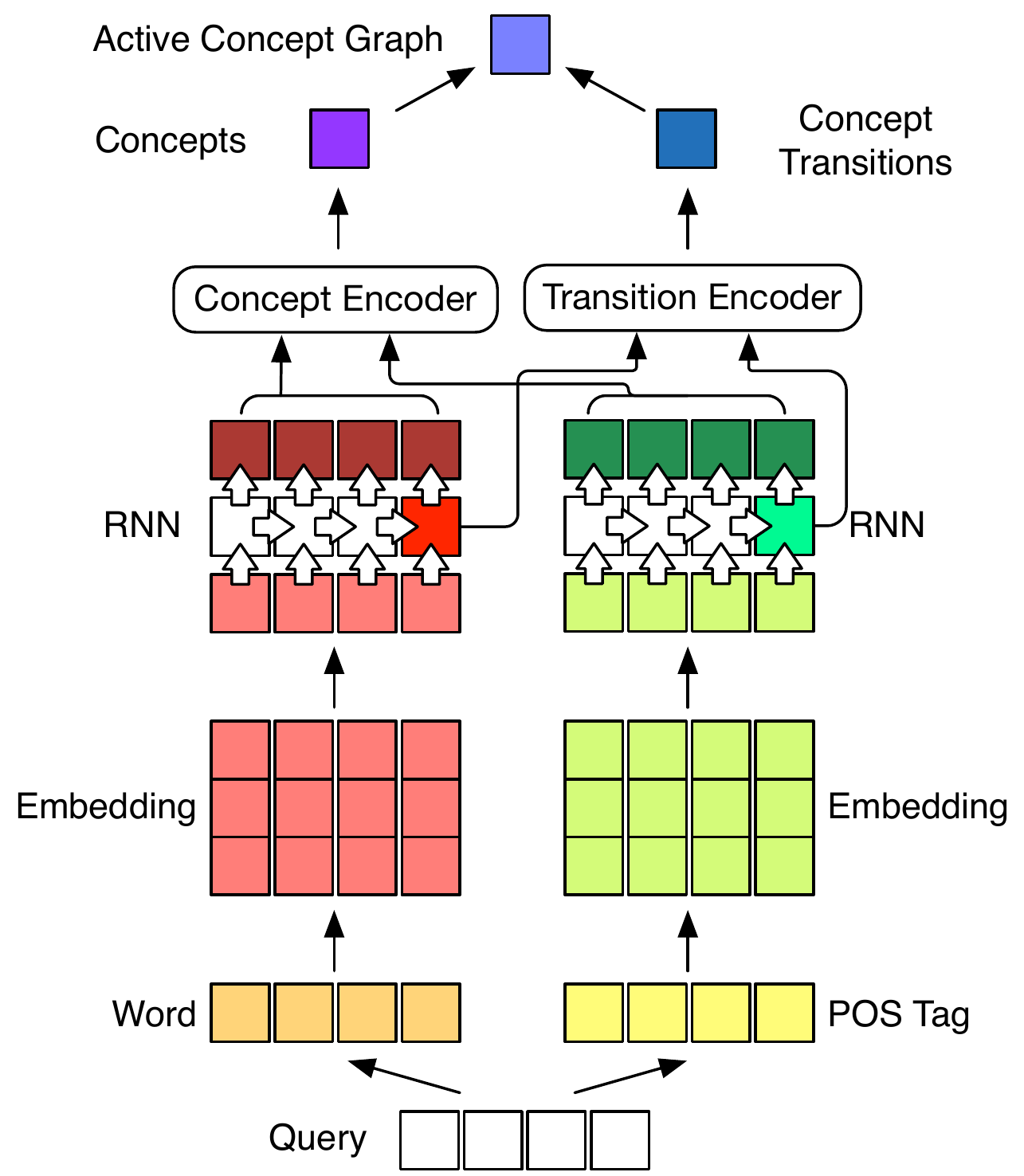, width=3.33in}
\caption{The proposed neural network architecture.}\label{fig::architecture}
\end{figure}
We introduce a novel neural network structure which provides an end-to-end solution to the concept transition inference problem where the input is a text query and the output is an active concept graph inferred by the given query. The model utilizes distributed representations for sequences of words and their POS tags respectively, from which semantic and syntax are embedded on a word-level. After that, two recurrent neural networks are adopted to model the sequential information from distributed representations of word and POS tag sequences in each query respectively. 
In the graph-based co-inference procedure, concepts and concept transition are inferred collectively and simultaneously. A concept encoder is proposed to utilize the joint outputs of two RNNs to encode each element into a concept vector.
% , which is later used to infer a probability distribution on all the concepts. 
Especially, the concept encoder is able to learn a confidence score which indicates the contribution of each element in encoding concept mentions in a query. While for inferring concept transitions, a transition encoder exploits the last hidden states of two RNNs to construct a transition vector, from which we infer a probability distribution on all possible concept transitions.
The loss of the neural network structure not only incorporates prediction errors between the predict concept transitions and the true concept transitions but also exploit a mutual transfer loss indicating the conflicts between the inferred concepts and their corresponding concept transitions. An active concept graph is presented with the inferred concepts and concepts transitions, by collectively minimizing a graph-based mutual transfer loss based on the concept graph. Figure \ref{fig::architecture} gives an overview of the proposed neural network architecture.

\subsection{Semantic-Syntax Representations}
Unlike traditional methods which ignore the sequential information of the input text query and treat it as a bag-of-words (BoW), in this work a text query $Q$ is considered as a sequence of elements $\{q_1, q_2, ..., q_K\}$, where each element $q_k$ can be a word or a phrase. $K$ is the length of a text query, which varies from different text queries. For each element $q_k$ in a text query $Q$, we utilize both the word indicating the semantic information, as well as its corresponding Part-of-Speech (POS) tag as the syntax information.

Part-of-speech (POS) tags bring useful syntax information about general word categories (such as noun, verb, adjective, etc.), which is helpful in dealing with ambiguous words and diversified expressions. For example, ``fever'' can be either a noun or a verb. The word ``fever'' with a POS tag ``noun'' is defined as a disease that causes an increase in body temperature and the fever with a POS tag ``verb'' can be considered as someone in a fever, as a symptom. In this work, an existing POS tagger\footnote{https://github.com/fxsjy/jieba} is utilized to give general POS tags to each element in the query. The semantic-syntax joint representation consists of words along with POS tags are shown to be effective in modeling both semantic (words) and syntax (POS tags) from the natural language text corpus in various tasks \cite{legrand2015joint,zhang2016mining}. In this work, each element $Q_k$ of a query $Q$ is represented by words and POS tags as a tuple:
\begin{equation}\label{eq::joint_feature}
{q_k} = \left( {{w_k},{p_k}}\right)~s.t.~w_k \in \mathbb{R}^{V_{word}}, p_k \in \mathbb{R}^{V_{pos}},
\end{equation}
where $w_k$ is the one-hot representation of the $k$-th word in the query $Q$ and $V_{word}$ is the number of unique words, namely the vocabulary size. Similarly, $p_k$ is the one-hot representation of the $k$-th word's POS tag in the query. $V_{POS}$ is the POS vocabulary size.

\subsection{Word Embedding}
The one-hot representation suffers from the curse of dimensionality since the representation becomes extremely sparse as the vocabulary becomes large. The word embedding is used to transfer one-hot representation of each word $w_k$ and POS tag $p_k$ into a dense representation:
\begin{equation}
{w\_{embed}}_{k} \in \mathbb{R}^{D_{word}}, {p\_{embed}}_{k} \in \mathbb{R}^{D_{pos}},
\end{equation} where $V_{word}$ usually can be large up to millions while $D_{word}$ is reduced to several hundreds. Note that ${D_{word}}$ and $D_{pos}$ are usually set empirically. In this work, we set ${D_{word}}=100$ and $D_{pos}=20$.

The embedded representation of each $w_k$ and $p_k$ are learned respectively by a linear mapping via a skip-gram model \cite{mikolov2013distributed}:
\begin{equation}
\begin{aligned}
{{embed}\_w}_{k} &= E_{word}~w_{k}\\
{{embed}\_p}_{k} &= E_{pos}~p_{k},
\end{aligned}
\end{equation}
where ${E_{word}} \in \mathbb{R}^{D_{word}\times V_{word}}$ and ${E_{pos}} \in \mathbb{R}^{D_{pos}\times V_{pos}}$ are weights.
The skip-gram learns a distributed representation of each word or POS tag based on its context. In the medical text queries, that means an explicit mention of a concept  (``Tylenol'') and an implicit mention of a concept  
(``Which medicine'') may have similar representations when they occur in similar context, when trained properly. That helps us solve the diversified expressions in medical text queries. 

In this work, the embedding is initialized with word vectors pre-trained from 64 million medical text queries and updates with the model during training.
After the word embedding, the $k$-th element in the text query $q_k$ has a semantic-syntax representation, represented by a tuple:
\begin{equation}
{e}_k = ({embed\_w}_k, {embed\_p}_k).
\end{equation}
% The word embedding:\\
% 1) reduce the dimension, which regularizes the model size\\
% 2) skip-gram learned by the context of text. help deal with real concept and vague concepts if they occur in similar context.

\subsection{Recurrent Neural Network}
Once we obtained semantic-syntax representations $e_k$ for each element $q_k$ in a query $Q$, the ${embed\_w}_k$ sequence and the ${embed\_p}_k$ sequences are fed into two recurrent neural networks, namely $\operatorname{RNN_W}$ and $\operatorname{RNN_P}$, respectively. 

In general, a recurrent neural network keeps hidden states over a sequence of elements and update the hidden state $h_k$ by the current input $x_k$ as well as the previous hidden state $h_{k-1}$ where $k>1$ by a recurrent function:
\begin{equation}
h_k = \operatorname{RNN}({x}_k, h_{k-1})
\end{equation}
The simplest form of an $\operatorname{RNN}$ is as follows:
\begin{equation}
h_k = \alpha(W_{xh}x_k+W_{hh}h_{k-1}+b_h),
\end{equation}
where $W_{xh} \in \mathbb{R}^{D_h\times D_{x}}, W_{hh} \in \mathbb{R}^{D_h \times D_h}, b_h \in \mathbb{R}^{D_h}$ are weights and bias that need to be learned as model parameters. $\alpha(\cdot)$ is a non-linear transformation function such as Rectified Linear Unit (ReLU): $\alpha(x)=max(0,x)$). This form of RNN fails to learn long-term dependencies due to gradient vanish or explosion problem \cite{bengio1994learning,hochreiter1998vanishing}, which is not suitable to learn dependencies from a long input sequence in practical.

% \subsubsection{GRU Cell}
To address the gradients decay or exploding problem over long sequences, the Gated Recurrent Unit (GRU) \cite{cho2014learning} is proposed as a variation of the Long Short-term Memory (LSTM) unit \cite{hochreiter1997long}. The GRU has been attracting great attentions since it overcomes the vanishing gradient in traditional RNNs and is more efficient than LSTM on certain tasks \cite{chung2014empirical}. The GRU is designed to learn from previous time stamps with long time lags of unknown size between important time stamps. A typical GRU is formulated as:
\begin{equation}
\begin{aligned}
  {r_k} &= \delta ({W_{xr}}{x_k} + {R_{hr}}{h_{k - 1}} + {b_r}) \hfill \\
  {z_k} &= \delta ({W_{xz}}{x_k} + {R_{hz}}{h_{k - 1}} + {b_z}) \hfill \\
  {{\tilde h}_k} &= tanh({W_{xh}}{x_k} + W_{hh}({r_k} \otimes {h_{k - 1}}) + {b_h}) \hfill \\
  {h_k} &= {z_k} \otimes {h_{k - 1}} + (1 - {z_k}) \otimes {{\tilde h}_k} \hfill ,\\
\end{aligned}
\end{equation}
where a reset gate $r_k$ is designed to makes the GRU acts whether as if it is reading the first element of an input sequence or not, allowing it to forget the previously computed state. The GRU maintains an update gate $z_k$ to balance between previous activation $h_{k-1}$ and the candidate activation ${{\tilde h}_k}$. $\delta(\cdot)$ and tanh$(\cdot)$ are the sigmoid and tangent activation function and $\otimes$ denotes the element-wise multiplication operator. An output vector $o_k$ is generated for each hidden state at time stamp $k$, by the following equation:
\begin{equation}
{o_k} = \sigma(W_{ho}{h_k})
\end{equation}
, where $W_{ho}$ is the weight and $\sigma(\cdot)$ is a softmax function. The output vector can be considered as the vector representation of each input $x_k$, taking the hidden state $h_k$ maintained by the RNN into the consideration. Note that, the output vector is not affected by any gates in GRU, which makes the GRU more appealing to our problem setting since we need an output vector $o_k$ without any output gating for each word in a query.

In this work, two separate RNN with GRU cells, namely $\operatorname{RNN}_W$ and $\operatorname{RNN}_P$, are adopted to model the sequential information for the sequence of embedded words ${embed\_w}_k$ and the sequence of embedded POS tags ${embed\_p}_k$:
\begin{equation}
\begin{aligned}
{h\_w}_k, {o\_w}_k &= \operatorname{RNN_W}({embed\_w}_k, {h\_w}_{k-1})\\
{h\_p}_k, {o\_p}_k &= \operatorname{RNN_P}({embed\_p}_k, {h\_p}_{k-1}),
\end{aligned}
\end{equation}

% \subsubsection{Bi-directional RNN}
% The usual RNN reads the sequence in an order from the first word to the last one: the hidden state at the time stamp summaries the current input as well as the proceeding hidden states. While, sometimes it is beneficial to take the following hidden states after the current hidden state into consideration, which results in a bi-directional RNN (Bi-RNN) \cite{schuster1997bidirectional}. A Bi-RNN consists of forward and backward RNNs: 
% \begin{equation}
% \overrightarrow{\operatorname{RNN}}(x_m, \overrightarrow{h}_{m-1}) ~~
% \overleftarrow{\operatorname{RNN}}(x_m, \overleftarrow{h}_{m+1}),
% \end{equation}
% where the hidden states $h_m$ of a Bi-RNN is the concatenation of hidden states of the forward direction and the backward direction $concat(\overrightarrow{{h_m}}, \overleftarrow {{h_m}})$ which are maintained separately.

\subsection{Graph-based Co-inference}
In order to fully exploit the correlations of concept transitions and corresponding concepts, concepts and concept transitions are inferred collectively over a concept graph for each query. The concept inference is aimed to select a subset of concept $\hat{C}_Q \in C$ that are mentioned in a query $Q$, which is achieved by the concept encoder. To inference transitions, we also utilize a transition encoder. The concepts $\hat{C}_Q$ and transitions $\hat{T}_Q$ are inferred collectively, by minimizing a mutual transfer loss which indicates the conflicts within the collectively inferred active concept graph $\hat{G}_Q$ on a concept graph $G$.
\subsubsection{Concept Encoder}%{\color{red}RNN with Attention}
In concept inference, a concept encoder is proposed to encode all the concept mentions from a sequence of output states of an $\operatorname{RNN}$ to concept vectors accordingly. Since some words in a query may contribute more to a concept mention in a query while some other words are less contributive, the concept encoder itself learns to assign a confidence score to each output state.
Let ${o}_k$ be the $k$-th output vector of an $\operatorname{RNN}$, while in this work we concatenate the output vectors of $\operatorname{RNN}_W$ and $\operatorname{RNN}_P$:
\begin{equation}
{o}_k = [{o\_w}_k, {o\_p}_k], {o\_w}_k\in\mathbb{R}^{1\times{D_{o_w}}},{o\_p}_k\in\mathbb{R}^{1\times{D_{o_p}}},
\end{equation} where $D_{o_w}$ and $D_{o_p}$ are the output dimension of output vectors in $\operatorname{RNN}_W$ and $\operatorname{RNN}_P$.
The concept encoder assigns a score $s_k$ for each $o_k$ indicating the degree of confidence based on the value of ${o}_k$:
\begin{equation}
{s_k} = CE({o_k},\theta )\quad {\text{s.t. }}\sum\nolimits_k {{s_k} = 1} ,\forall {s_k} \in [0,1],
\end{equation}
where $\theta$ is the parameter of the concept encoder that we learn along with the whole model. all $s_k$ scores in a query are normalized to sum up to one. The concept encoder $CE$ can be also considered as a mapping from each output vector $o_k$ to a real value $s_k\in[0,1]$. In this work, the concept encoder is implemented as a single layer neural network with a non-linear activation function ReLU. Thus $\theta=\{W_{\theta}\in \mathbb{R}^{(D_{ow}+D_{op})\times1}, b_{\theta}\in{\mathbb{R}}\}$. Note that although weights and biases are applied on each of the $o_k$, they are shared among all ${o_1,o_2,...,o_k}$.

\begin{figure}[hbtp]
\centering
\epsfig{file=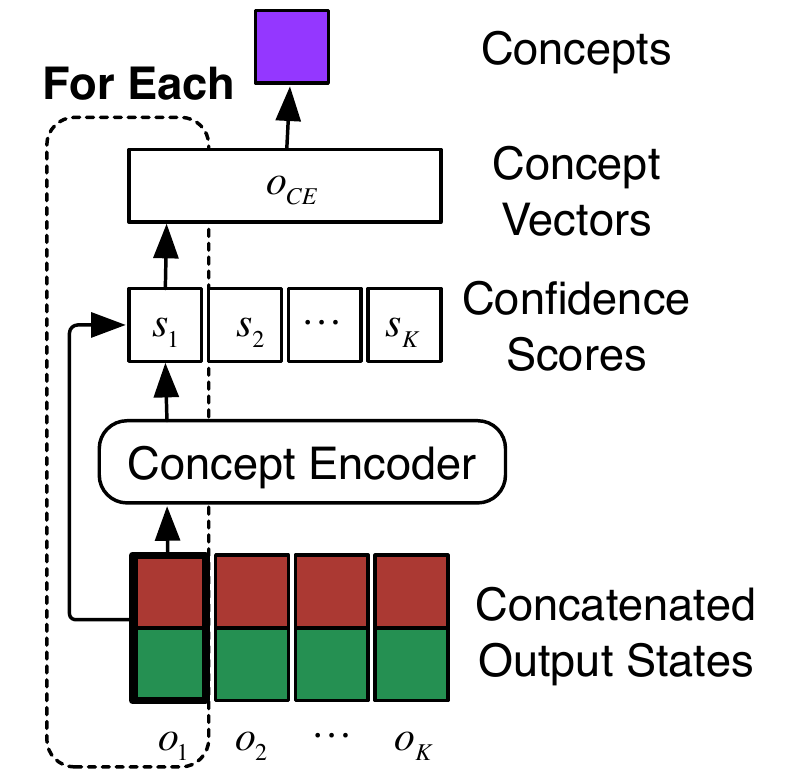, height=2in}
\caption{The concept encoder is used to determine confidence scores for each joint output state. This figure shows an example of a score $s_1$ learned from the concept encoder for $o_1$.}\label{fig::concept_encoder}
\end{figure}

% The concept encoder module assigns an attentive value for each output state of the RNN on a character level, which is aimed to determine the contribution of each word to the activation of a concept.
% \begin{equation}
% {o_{AN}} = \operatorname{AN} (o) = \sum\limits_{i = 1}^M {score({o_i}) \cdot {\kern 1pt} } {o_i},\quad {\text{s.t. }}\sum\nolimits_i {score({o_i}) = 1} ,\forall score({o_i}) \in [0,1]
% \end{equation}
Figure \ref{fig::concept_encoder} shows the architecture of the concept encoder.
The $o_{CE}\in\mathbb{R}^{(D_{ow}+D_{op})\times{K}}$ is a representation of encoded concepts from the query, which is calculated based on all the output states $\{o_1, o_2,...,o_K\}$:
\begin{equation}
O_{CE}=\left[ {\begin{array}{*{20}{c}}
  {{{({CE(o_1,\theta)} \cdot {o_1})}^T}}&{...}&{{{({CE(o_K,\theta)} \cdot {o_K})}^T}}
\end{array}} \right]
\end{equation}

The probability that a concept $c_i\in{C}$ is activated in a query $Q$ is defined as:
\begin{equation}
({\hat{C}_{Q}})_m = P(c_m|c_m\in{C},\theta,W_{CE},b_{CE})=\frac{1}{{1 + {e^{ - {W_{CE}}{O_{CE}} + {b_{CE}}}}}},
\end{equation}
where $W_{CE} \in \mathbb{R}^{1 \times (D_{ow}+D_{op})}$, $b_{CE} \in \mathbb{R}$ are weights and biases for such probability inference. 
% For varying query length $K$ in different queries, zero padding is used for the end of each query. 
We use $\hat{C}_Q\in\mathbb{R}^{1\times{M}}$ to quantify the probability distribution on all $M$ concepts for a given query $Q$. 
% The correlation of the concepts being predicted with the attentive values set on a character level. Need to have a case study in the experiment section.

\subsubsection{Transition Encoder}
In the field of machine translation, a novel recurrent neural network encoder-decoder has gained attention \cite{sutskever2014sequence}, where the encoder recurrent neural network encodes the global information spanning over the whole input sentence in its last hidden state. The effectiveness of the last hidden states in modeling natural language sequences are also witnessed in application like dialog systems \cite{serban2016building}. Inspired by those ideas, we propose a transition encoder which leverages the last hidden state of the neural network for both $\operatorname{RNN}_W$, $\operatorname{RNN}_P$ to make inferences on concept transitions, where the transition vector $o_{TE}$ is constructed by
% 1) further exploited the discrepancies as well as similarities among different outputs by multiple feature maps. \\
% 2) the concept transition 
\begin{equation}
O_{TE} = [{h\_w}_K,{h\_p}_K],
\end{equation}
where $K$ is the length of the query.
The probability that a transition $t_{n}\in{T}$ is activated given a query $Q$ is quantified by:
\begin{equation}
(\hat{T}_{Q})_{n}=P(t_{n}\in{T}|W_{TE},b_{TE})=\frac{1}{{1 + {e^{ - {W_{TE}}{O_{TE}} + {b_{TE}}}}}},
\end{equation}
where $W_{TE} \in \mathbb{R}^{1 \times (D_{ow}+D_{op})}, b_{TE} \in \mathbb{R}$ are weights and biases for the transition encoder. Similarly, $\hat{T}_Q\in\mathbb{R}^{1\times{N}}$ denotes the inferred probability distribution on all $N$ concept transitions given a query $Q$. 

\subsection{Mutual Transition Loss}
The idea of mutual transition loss is to characterize the loss caused by transferring the inferred concept transitions to their corresponding concepts, and the other way around. Since for each concept transition $t_{{i}\to{j}}\in{T}$, two concepts $c_i$ and $c_j$ are evolved in the query. If a concept transition $t_{{i}\to{j}}$ is inferred with a high probability while its corresponding concepts $c_i$, $c_j$ have low probabilities, then that indicates conflicts in the final active concept graph. The mutual transition loss is proposed in the co-inference procedure to minimize the conflicts between the inferred concepts and concept transitions so that the resulting active concept graph can be more reasonable.

The graph-based formulation for concept graph gives an appealing property that transitions and their proximate concepts can be clearly characterized by a transfer matrix $A\in\mathbb{R}^{M\times{N}}$ over the concept graph $G=\langle{C},{T}\rangle$. Each entry $a_{mn}=1$ if and only if the concept $c_m$ involves in at least a concept transition $t_{m\to{\cdot}}$ or $t_{{\cdot}\to{m}}$.
% \begin{equation}
% a_{mn}=1~\Leftrightarrow~\exists~t_{m\to{c}}| c\in C
% \end{equation}

The mutual transfer loss is defined on $\hat{C}_Q, \hat{T}_Q, \mathcal{T}_Q$ as:
\begin{equation}
\mathcal{L}_{MTL}(\hat{C}_Q, \hat{T}_Q, \mathcal{T}_Q) = H(\mathcal{T}_Q, \hat{T}_Q) + E(\hat{C}_Q, \hat{T}_Q),
\end{equation}
where $\mathcal{T}_Q$ is a ground truth one-hot indicator for concept transitions given a query $Q$. $\hat{C}_Q$ and $\hat{T}_Q$ are inferred concepts and concept transitions with the proposed method. $H(\cdot, \cdot)$ calculates the cross entropy \cite{tsoumakas2009mining}. $E(\hat{C}_Q, \hat{T}_Q)$ is an energy-based function on inferred transitions $\hat{T}_Q$ and inferred concepts $\hat{C}_Q$. Each combination of $\hat{C}_Q$ and $\hat{T}_Q$ corresponds with an energy value, the lower energy level a combination of $\hat{C}_Q$ and $\hat{T}_Q$ has indicates less conflicts among the inferred concepts and transitions. In this work, an energy-based function for $E(\hat{C}_Q, \hat{T}_Q)$ is proposed as:
\begin{equation}
 E(\hat{C}_Q, \hat{T}_Q) = \mathcal{L}_R(\hat{C}_Q, \hat{T}_Q{{A}^T}) + \mathcal{L}_R(\hat{T}_Q,\hat{C}_Q{A}),
\end{equation}
where $\mathcal{L}_R$ is similar with the ranking loss \cite{murphy2012machine}. In this work, $\mathcal{L}_R$ penalizes cases where the inferred concepts/transitions after transformation by matrix $A$ have high probabilities but order below the ranking of the originally inferred concepts/transitions in a query. $\mathcal{L}_R$ has a general form:
\begin{equation}
\begin{gathered}
%   \mathcal{L}(X,\hat X) = \frac{1}{N}\sum\limits_{i = 1}^N {\frac{1}{{\left| {{{\hat X}_i}} \right|(M - \left| {{{\hat X}_i}} \right|)}}\left| {} \right.}  \hfill \\
%   \quad \quad \{ \left( {p,q} \right):{{\hat X}_{ip}} < {{\hat X}_{iq}},{X_{ip}} = 1,{X_{iq}} = 0\}  \hfill \\ 
  \mathcal{L}_{R}(\hat X,\hat Y) = {\frac{1}{{\left| {{{\hat X}}} \right|(L - \left| {{{\hat X}}} \right|)}}\left| {} \right.} 
\{ \left( {p,q} \right):{{\hat Y}_{p}} < {{\hat Y}_{q}},{\hat X_{p}} \geq{\hat X_{q}} \},  \hfill \\
\end{gathered}
\end{equation}
where $\hat X\in{\mathbb{R}^{1\times{L}}}$ is the originally inferred labels and $\hat Y\in{\mathbb{R}^{1\times{L}}}$ is the inferred labels from the transformation with $A$. $\left|\cdot\right|$ denotes the number of ground truth labels being assigned. $L$ is the label size, where we have $M$ for concepts and $N$ for concept transitions.

% \begin{equation}
% E(C',T') = \alpha L(C',T'A) + \beta L(T',T'{A^T})
% % E(CT',C') = \left\| {T'P - C'} \right\|_2 + \mu\left\| {C'P^{T}-T'} \right\|_2
% \end{equation} is used in this paper, where $\left\|.\right\|_2$ denotes the $\ell_2$ norm and $P \in \mathbb{R}^{X \times Y}$ is the transfer matrix.
% Since in this work, we consider concepts and concept transitions are equally .. therefore $\mu=1$.

%% file: subfiles/4_experiments.tex
\section{Evaluation}
\subsection{Data Set}\label{sec::data}
We collect medical queries from an online medical question answering forum\footnote{http://club.xywy.com}, on which user posted their healthcare related questions and medical professionals give online suggestions or advice. 
% Questions from patients with concept transitions are selected.
% How to annotate 1. seed 2. crowdsourced 3. majority voting
% Medical professionals are contacted to review a seed of annotated concept mentions and concept transitions, on 1000 real-world medical text queries. 
% Then data is annotated on a crowdsourcing platform\footnote{http://test.baidu.com}. Each annotator in the crowd will be presented with a yes/no question like ``Does the following question contains the concept transition \textit{Symptom}$\to$\textit{Medicine}$\to$\textit{Instruction}?'' or ``Does the following question mentions the medicine concept'', followed by a seed question which does have such concept transition/concept mention. The majority voting from five annotators are used as the ground truth label for each text queries. 
% Table \ref{tab::query_stats} shows the statistics on collected medical text queries. 
% space
% \begin{table}[hbt]
% \centering
% \begin{tabular}{llll}
% \toprule
% Vocabulary         & Length         & $\left|\bar{C}\right|$ & $\left|\bar{T}\right|$\\ \hline
% 11521 (word) & \multirow{2}{*}{13.8291$\pm$6.0429} & \multirow{2}{*}{3.6020} & \multirow{2}{*}{2.4723} \\
% 60 (POS)       &                                     &        &   \\
% \bottomrule
% \end{tabular}
% \caption{Statistics on collected medical queries.}\label{tab::query_stats}
% \end{table}
The obtained corpora are in Chinese. Due to the fact that Chinese text queries are not naturally split by spaces, word segmentation is performed using a Chinese word segmentation package \cite{jieba}. The segmentation results do not simply segment queries by each Chinese character. Instead, it tries to combine strongly correlated consecutive characters into words, thus “word” referred in this work can contain more than one Chinese character.
After preprocessing and annotation, a medical text query has the following format: \{
``text'': ``\begin{CJK*}{UTF8}{gbsn}宫颈管~慢性~炎症~伴~鳞状~上皮~内~挖空~细胞~聚集~是~宫颈癌~吗~严重~吗~需要~leep~手术~吗\end{CJK*}'', ``pos'': ``n~b~n~v~n~n~n~n~n~v~v~n~y~a~y~v~eng~n~y'', ``concept'': ``fee$|$disease$|$surgery$|$recover$|$treatment'', ``concept\_transition'': ``disease $\to$ surgery $\to$ recover''\}, where the POS tagging uses ICTCLAS annotation \cite{zhang2003hhmm}. Among 10,000 medical text queries, 11,531 unique words and 60 unique POS tags are observed. The average length of question is 13.8, with a standard variation of $\pm$6.1. The average number of concepts in labeled queries is 3.6020$\pm$0.8. The average number of concept transitions is 2.4723$\pm$0.7.
Word embeddings are pre-trained using a skip-gram model \cite{mikolov2013distributed} on 64 million unlabeled medical text queries separately.
% By a separately trained word embedding model using large corpus in a totally unsupervised fashion, we can alleviate the negative impact from limited word embedding training corpus from only labeled queries.
Context window size is set to 8 and we specify a minimum occurrence count of 5. The vocabulary contains 100-dimension vectors on 382216 words. Words not presented in the set of pre-trained words are initialized as random vectors. All word vectors will be updated during training.

\subsection{Experiment Settings}
\subsubsection{Comparison Methods}
To show the advantages of the proposed method in addressing the concept transition inference problem, we compare it with the following baseline models. 
\begin{itemize}
\item LR: a logistic regression model applied with POS tagging features and word representations.\item NNID-JM \cite{zhang2016mining}: the neural network intention detection model with joint modeling. Both words and POS tags are used to characterize the question . Domain-specific POS tags, such as ``noun\_medicine'', are used in NNID-JM instead of ``noun'' for word ``Tylenol''. The NNID-JM doesn't explicitly exploit label correlations on the output level.
\item CI: the concept inference model which only infers mention of concepts from queries with the concept encoder. $H(\mathcal{C}_Q, \hat{C}_Q)$ is used as the loss function for the CI task.
\item CTI: the concept transition inference model without co-inference. Only concept transitions are inferred from queries without considering concepts. The last output states of two RNNs are concatenated to predict the concept transitions. $H(\mathcal{T}_Q, \hat{T}_Q)$ is used as the loss function.
\item coCTI: the concept transition inference model with co-inference. $H(\mathcal{T}_Q, \hat{T}_Q) + H(\mathcal{C}_Q, \hat{C}_Q)$ is used as the loss function. This variation can be seen as a multi-task learning model for concept and concept transitions, where both tasks share the neural network structure for word representation.
\item coCTI-MTL: the proposed model with co-inference and a mutual transfer loss $\mathcal{L}_{MTL}$, where the CI task and CTI task not only share the neural network structure, but also guided by the mutual transfer loss.
\end{itemize}

\subsubsection{Evaluation Metrics}
Each edge in the concept graph is considered as an individual label and we evaluate inferred concept transitions as a multi-class, multi-label classification problem. 
Receiver operating characteristic (ROC) \cite{hanley1982meaning}, the micro/macro-average area under the curve (micro-AUC, macro-AUC) \cite{cortes2004auc}, coverage error \cite{tsoumakas2009mining} and label ranking average precision (LRAP) \cite{madjarov2012extensive} are used to evaluate the effectiveness of the proposed model in inferring concept transition in medical text queries. The ROC and AUCs focus on the quality of prediction, while the coverage error and LRAP are introduced to evaluate the completeness/ranking of the prediction. ROC is the curve created by plotting the true positive rate (TPR) against the false positive rate (FPR) at various threshold settings. Micro-AUC computes the averaged area under the ROC curve over all the labels. Coverage error computes the average number of labels that we need to have in the final prediction in order to predict all true labels. LRAP score favors better rank to the labels associated to each sample, which is used in multi-label ranking problems.

\subsubsection{Experiment Settings}
The embeddings for word and POS tagging have a dimension of 100 and 20, respectively. The hidden layer and the output layer of the GRU unit have a dimension of 100. For training the proposed neural network structure, 70\% of the labeled data are used for training and 10\% data serve as a validation set to tune for the best parameter set. The remaining data are used for testing. Cross-validation is used and we combine test data in each fold to report the test performance.
The optimization is performed in a mini-batch fashion with a batch size of 32. The Adam Optimizer \cite{kingma2014adam} is applied to train the neural network and the initial learning rate is set to $10^{-4}$. Weight variables are initialized with the Xavier initializer \cite{glorot2010understanding} and bias variables are initialized as zeros.
The proposed model is implemented in Tensorflow \cite{abadi2015tensorflow}.

\subsection{Evaluation Results}
\begin{figure}[h]
\centering
\epsfig{file=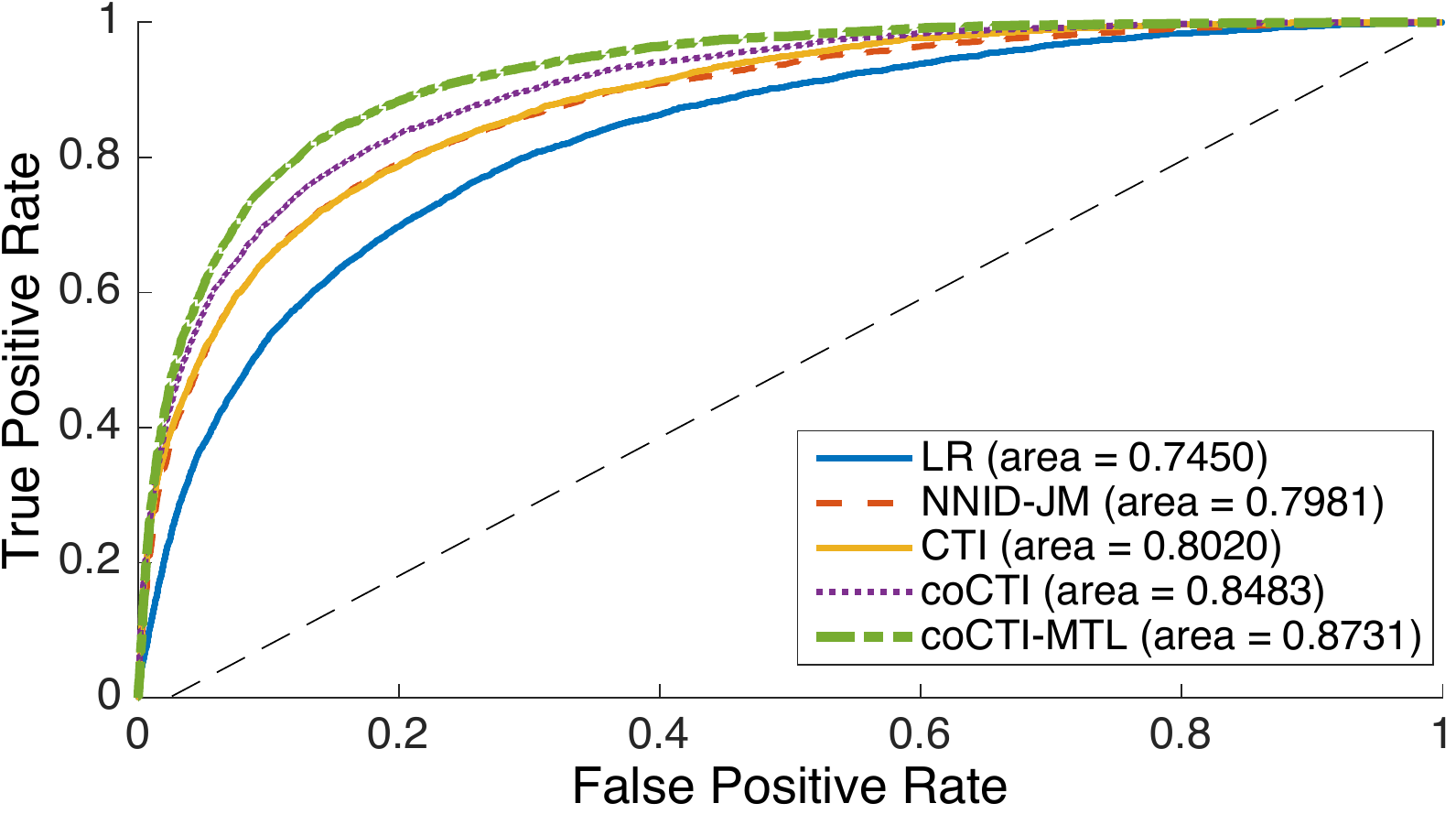, width=3.3in}
\caption{micro-AUC scores and ROC curves.}\label{fig::mroc}
\end{figure}
Figure \ref{fig::mroc} shows the effectiveness of the proposed model by micro-AUC and ROC curves. Generally, neural network based models (NNID-JM, CTI, coCTI, coCTI-MTL) outperform traditional logistic regression model (LR) consistently. 
For NNID-JM, in order to make a fair comparison, domain specific POS tags (such as noun\_disease, noun\_medicine, noun\_symptom) are maintained as an external knowledge base. Those POS tags are used by the POS tagger in NNID-JM as its default setting. When compared with NNID-JM, the proposed CTI model achieves similar performance on micro-AUC, while it doesn't rely on any other external knowledge like domain-specific POS tags in NNID-JM. In practical, utilizing a concept transition graph is usually more feasible than tagging words and building dictionaries to maintain words for each domain-specific concept.

From Figure \ref{fig::mroc} we can further observe that CTI-MTL achieves the best performance (0.8731 in micro-AUC) among all the comparison methods in inferring concept transitions in medical queries.  The CTI-MTL model has a nearly 2.5\% improvement on micro-AUC when compared with coCTI and a nearly 7.5\% improvement with CTI. This demonstrates that the mutual transfer loss which penalizes conflicts between the inferred concepts and inferred concept transitions can improve the inference quality.

% \begin{table}
% \centering
% \begin{tabular}{llll}
% \toprule
% Method & CI & CTI & coCTI\\\hline
% Micro-AUC & 0.8217    & 0.8020    & 0.8483 \\
% \bottomrule
% \end{tabular}
% \caption{Micro-AUC scores for collective inference (coCTI) VS. inference on each task (CI/CTI) individually.}\label{tab::coinferornot}
% \end{table}

\begin{figure}[h]
\centering
\epsfig{file=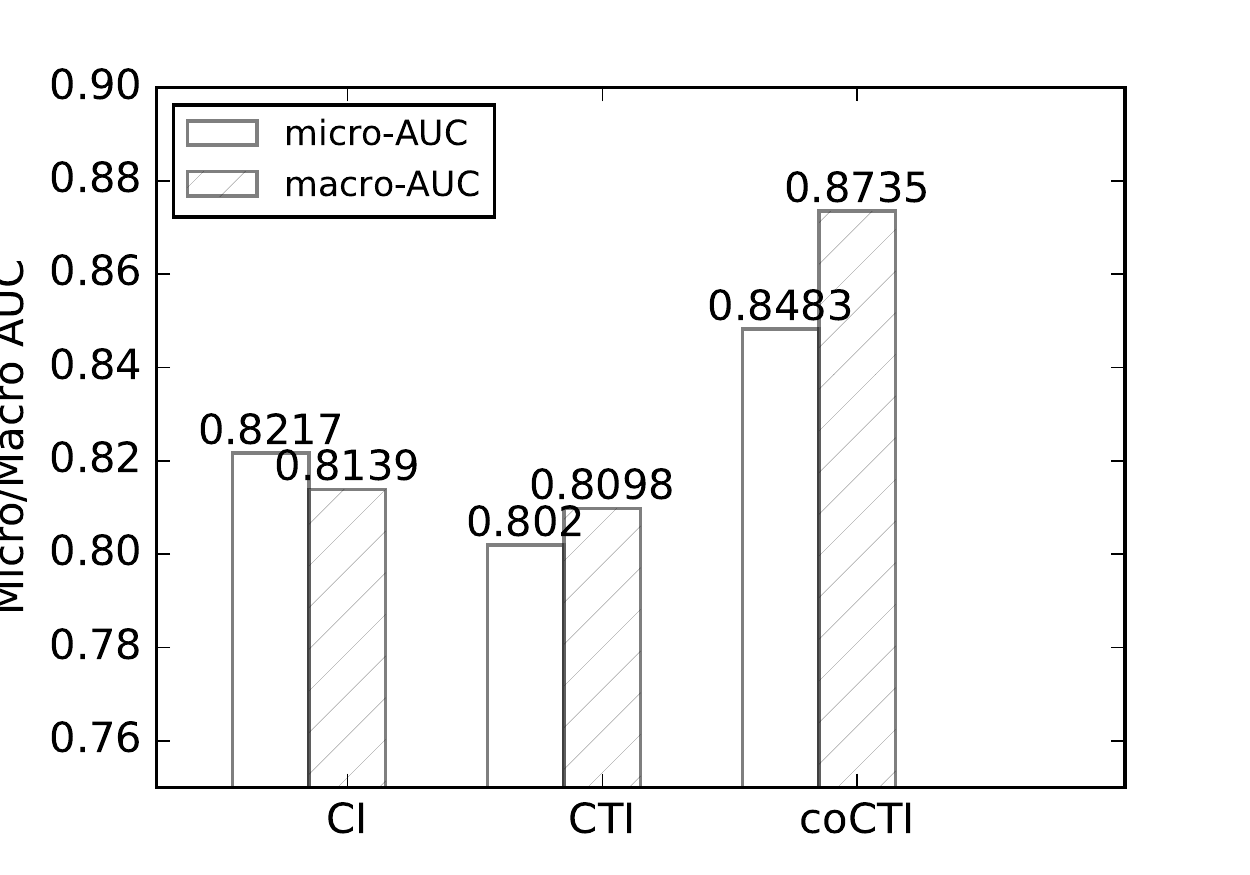, width=3.3in}
\caption{Micro/Macro-AUC scores for collective inference (coCTI) VS. concept inference(CI) and concept transition inference (CTI) separately.}\label{fig::coinferornot}
\end{figure}

% \begin{table}[htb]
% \centering
% \resizebox{\columnwidth}{!}{%
% \begin{tabular}{cccccc}
% \toprule
% Method        & LR          & NNID-JM     & CTI          & coCTI         & coCTI-MTL   \\\hline
% Coverage Loss & 8.2290 & 7.1586 & 7.4013 & 6.5874 & 5.4794\\
% LRAP &&&&&\\\bottomrule
% \end{tabular}}
% \caption{Coverage Loss}\label{tab::cl}
% \end{table}

\begin{table}[h]
\centering
\resizebox{\columnwidth}{!}{%
\begin{tabular}{lccccc}
\toprule
Concept Transition & LR& NNID-JM& CTI& coCTI & coCTI-MTL\\ \hline
\textit{Symptom}$\to$\textit{Diet} & 0.6544 {\color{blue}(5)}& 0.7755 {\color{blue}(4)}& 0.7669 {\color{blue}(3)}& 0.7959 {\color{blue}(2)}& 0.8495 {\color{blue}(1)}\\
\textit{Symptom}$\to$\textit{Medicine} & 0.7022 {\color{blue}(5)} & 0.7893 {\color{blue}(4)} & 0.8242 {\color{blue}(3)} & 0.8571 {\color{blue}(2)} & 0.8624 {\color{blue}(1)}\\
\textit{Symptom}$\to$\textit{Cause}  & 0.7600 {\color{blue}(5)} & 0.8549 {\color{blue}(4)} & 0.8786 {\color{blue}(3)} & 0.8911 {\color{blue}(1)} & 0.8880 {\color{blue}(2)}\\
\textit{Disease}$\to$\textit{Diet}  & 0.7818 {\color{blue}(5)} & 0.8670 {\color{blue}(4)} & 0.8681 {\color{blue}(3)} & 0.9059 {\color{blue}(2)}& 0.9458 {\color{blue}(1)}\\
\textit{Disease}$\to$\textit{Treatment}  & 0.7181 {\color{blue}(5)} & 0.7787 {\color{blue}(3)} & 0.7482 {\color{blue}(4)} & 0.8456 {\color{blue}(2)}& 0.8836 {\color{blue}(1)}\\
\textit{Disease}$\to$\textit{Examine}  & 0.6397 {\color{blue}(5)} & 0.6707 {\color{blue}(4)} & 0.7838 {\color{blue}(3)} & 0.8221 {\color{blue}(2)} & 0.8480 {\color{blue}(1)}\\
\textit{Disease}$\to$\textit{Medicine} & 0.7623 {\color{blue}(5)} & 0.8726 {\color{blue}(4)} & 0.8749 {\color{blue}(3)} & 0.8873 {\color{blue}(2)} & 0.9015 {\color{blue}(1)}\\
\textit{Surgery}$\to$\textit{Recover}  & 0.8117 {\color{blue}(5)} & 0.9126 {\color{blue}(3)} & 0.9012 {\color{blue}(4)} & 0.9239 {\color{blue}(2)} & 0.9396 {\color{blue}(1)}\\
\textit{Surgery}$\to$\textit{Sequela}  & 0.7385 {\color{blue}(5)} & 0.8031 {\color{blue}(4)} & 0.8214 {\color{blue}(3)} & 0.8417 {\color{blue}(2)} & 0.8972 {\color{blue}(1)}\\
\textit{Surgery}$\to$\textit{Syndrome}  & 0.7896 {\color{blue}(5)} & 0.7994 {\color{blue}(4)} & 0.8634 {\color{blue}(2)} & 0.8619 {\color{blue}(3)} & 0.9172 {\color{blue}(1)}\\
\textit{Surgery}$\to$\textit{Risk}  & 0.6613 {\color{blue}(5)} & 0.8063 {\color{blue}(4)} & 0.8688 {\color{blue}(3)} & 0.8715 {\color{blue}(2)} & 0.9099 {\color{blue}(1)}\\
\textit{Medicine}$\to$\textit{Symptom}  & 0.6861 {\color{blue}(5)} & 0.8275 {\color{blue}(3)} & 0.7553 {\color{blue}(4)} & 0.8294 {\color{blue}(2)} & 0.8598 {\color{blue}(1)}\\
\textit{Medicine}$\to$\textit{Side Effect}  & 0.6652 {\color{blue}(5)} & 0.8162 {\color{blue}(3)} & 0.7771 {\color{blue}(4)} & 0.8135 {\color{blue}(2)} & 0.8814 {\color{blue}(1)}\\
\textit{Medicine}$\to$\textit{Disease}  & 0.6806 {\color{blue}(4)} & 0.6514 {\color{blue}(5)} & 0.8081 {\color{blue}(3)} & 0.8126 {\color{blue}(2)} & 0.8678 {\color{blue}(1)}\\
\textit{Medicine}$\to$\textit{Instruction}  & 0.7090 {\color{blue}(5)} & 0.7761 {\color{blue}(3)} & 0.7603 {\color{blue}(4)} & 0.8170 {\color{blue}(2)} & 0.8820 {\color{blue}(1)}\\
\textit{Examine}$\to$\textit{Fee}  & 0.7576 {\color{blue}(5)} & 0.9049 {\color{blue}(3)} & 0.8981 {\color{blue}(4)} & 0.9425 {\color{blue}(2)} & 0.9482 {\color{blue}(1)}\\
\textit{Examine}$\to$\textit{Diagnosis}  & 0.6832 {\color{blue}(5)} & 0.7956 {\color{blue}(3)} & 0.7445 {\color{blue}(4)} & 0.8383 {\color{blue}(2)} & 0.8822 {\color{blue}(1)}\\
\textit{Symptom}$\to$\textit{Treatment}  & 0.6817 {\color{blue}(5)} & 0.7640 {\color{blue}(3)} & 0.7313 {\color{blue}(4)} & 0.8130 {\color{blue}(2)} & 0.8531 {\color{blue}(1)}\\
\textit{Symptom}$\to$\textit{Department}  & 0.5978 {\color{blue}(5)} & 0.6460 {\color{blue}(3)} & 0.6013 {\color{blue}(4)} & 0.6738 {\color{blue}(2)} & 0.8080 {\color{blue}(1)}\\
\textit{Disease}$\to$\textit{Cause}  & 0.7306 {\color{blue}(5)} & 0.8206 {\color{blue}(4)} & 0.8515 {\color{blue}(3)} & 0.8608 {\color{blue}(2)} & 0.8634 {\color{blue}(1)}\\
\textit{Disease}$\to$\textit{Symptom}  & 0.6936 {\color{blue}(4)} & 0.7552 {\color{blue}(3)} & 0.6845 {\color{blue}(5)} & 0.7554 {\color{blue}(2)} & 0.8372 {\color{blue}(1)}\\
\textit{Disease}$\to$\textit{Department}  & 0.6931 {\color{blue}(5)} & 0.7387 {\color{blue}(4)} & 0.7431 {\color{blue}(3)} & 0.7652 {\color{blue}(2)} & 0.8290 {\color{blue}(1)}\\
\textit{Disease}$\to$\textit{Surgery}  & 0.7801 {\color{blue}(5)} & 0.8795 {\color{blue}(4)} & 0.9029 {\color{blue}(3)} & 0.9236 {\color{blue}(2)} & 0.9380 {\color{blue}(1)}\\\bottomrule
\end{tabular}
}
\caption{Fine-grained AUC scores for concept transition inference for each concept transition (each edge in the concept graph).}\label{tab::fineauc}
\end{table}

Figure \ref{fig::coinferornot} shows the effectiveness of the co-inference procedure by comparing the performance of CTI with coCTI. The CI infers concept mentions so we can't simply compare its performance with CTI/coCTI where concept transitions are inferred. However, for CTI and coCTI, the improved performance on both micro-AUC and macro-AUC validate the effectiveness of inferring concept transitions and concept mentions collectively than inferred separately. The coCTI model can be considered as a multi-task learning model where the question representation is learned jointly and shared between two inference tasks.

Furthermore, the fine-grained AUC scores on all concept transitions without micro/macro-averaging are shown in Table \ref{tab::fineauc}. A general observation we can draw from the results is that the coCTI-MTL model is able to outperform other baselines in almost all types of concept transitions. 

\begin{figure}[tbh!]
\centering
\epsfig{file=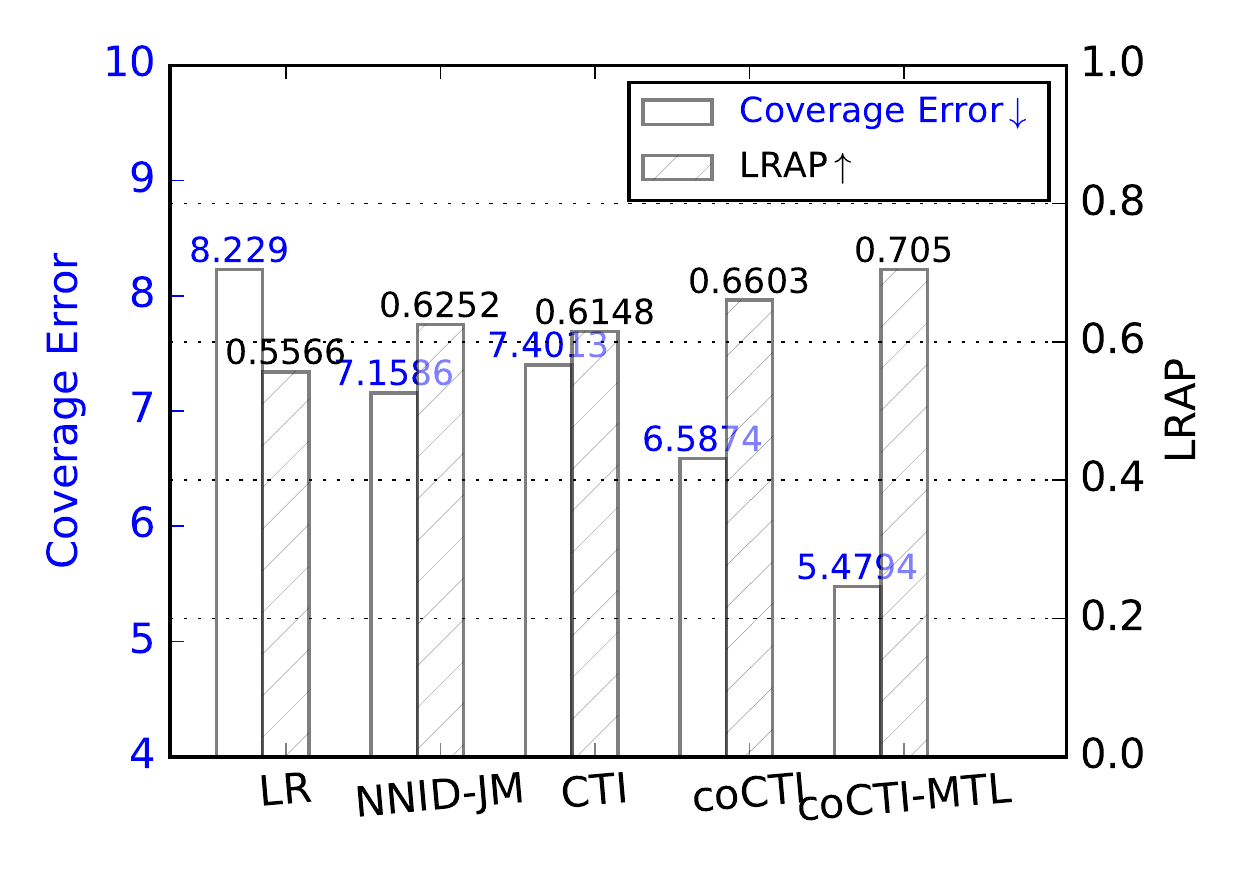, width=3.3in}
\caption{Coverage Loss and Label Ranking Average Precision (LRAP).}\label{fig::cl_lrap}
\end{figure}

Figure \ref{fig::cl_lrap} shows the coverage loss and LRAP over proposed methods and other baselines, where the coCTO-MTL model is able to achieve the lowest coverage error and the highest label ranking average precision score.
% \input{subfiles/4_experiments_pr.tex}

% \subsection{Case Study}
\input{subfiles/4_experiments_casestudy.tex}
Five case studies are presented in Figure \ref{fig::cases} to show scores assigned by the concept encoder on real-world medical text queries. Some stop-words are removed for clarity. We can see that the concept encoder is properly trained as it is able to assign important words or words refer to concepts higher confidence scores, while common words are less likely to receive such high scores. This observation indicates the effectiveness of the proposed concept encoder in encoding concept mentions without relying on domain-specific external knowledge bases.
% Effectiveness of the concept encoder\\
% show that concepts with concept transitions associate with larger attentive values than those inactive concepts.

%% file: subfiles/4_experiments_casestudy.tex
\begin{figure}[h]
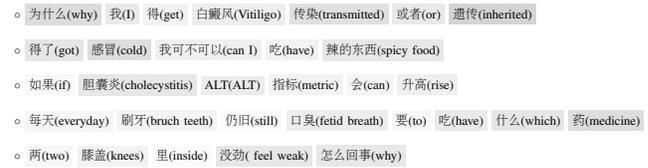

\centering
\begin{adjustbox}{width=\columnwidth}
\begin{tabular}{l}
% \toprule		
$\circ$~\colorbox{black!13.6991684}{\begin{CJK*}{UTF8}{gbsn}为什么\end{CJK*}(why)}~\colorbox{black!8.2321416}{\begin{CJK*}{UTF8}{gbsn}我\end{CJK*}(I)}~\colorbox{black!5.2720109}{\begin{CJK*}{UTF8}{gbsn}得\end{CJK*}(get)}~\colorbox{black!7.1961719}{\begin{CJK*}{UTF8}{gbsn}白癜风\end{CJK*}(Vitiligo)}~\colorbox{black!12.1971045}{\begin{CJK*}{UTF8}{gbsn}传染\end{CJK*}(transmitted)}~\colorbox{black!8.5471444}{\begin{CJK*}{UTF8}{gbsn}或者\end{CJK*}(or)}~\colorbox{black!16.9722869}{\begin{CJK*}{UTF8}{gbsn}遗传\end{CJK*}(inherited)}\\\\
$\circ$~\colorbox{black!9.9837363}{\begin{CJK*}{UTF8}{gbsn}得了\end{CJK*}(got)}~\colorbox{black!12.9141718}{\begin{CJK*}{UTF8}{gbsn}感冒\end{CJK*}(cold)}~\colorbox{black!5.8589015}{\begin{CJK*}{UTF8}{gbsn}我可不可以\end{CJK*}(can I)}~\colorbox{black!4.9853317}{\begin{CJK*}{UTF8}{gbsn}吃\end{CJK*}(have)}~\colorbox{black!8.8892005}{\begin{CJK*}{UTF8}{gbsn}辣的东西\end{CJK*}(spicy food)}\\\\
$\circ$~\colorbox{black!2.2837764}{\begin{CJK*}{UTF8}{gbsn}如果\end{CJK*}(if)}~\colorbox{black!9.2840053}{\begin{CJK*}{UTF8}{gbsn}胆囊炎\end{CJK*}(cholecystitis)}~\colorbox{black!7.5126059}{\begin{CJK*}{UTF8}{gbsn}ALT\end{CJK*}(ALT)}~\colorbox{black!5.0292235}{\begin{CJK*}{UTF8}{gbsn}指标\end{CJK*}(metric)}~\colorbox{black!3.5806343}{\begin{CJK*}{UTF8}{gbsn}会\end{CJK*}(can)}~\colorbox{black!6.3413799}{\begin{CJK*}{UTF8}{gbsn}升高\end{CJK*}(rise)}~\\\\
$\circ$~\colorbox{black!3.27}{\begin{CJK*}{UTF8}{gbsn}每天\end{CJK*}(everyday)}~\colorbox{black!4.26}{\begin{CJK*}{UTF8}{gbsn}刷牙\end{CJK*}(bruch teeth)}~\colorbox{black!1.86}{\begin{CJK*}{UTF8}{gbsn}仍旧\end{CJK*}(still)}~\colorbox{black!7.01}{\begin{CJK*}{UTF8}{gbsn}口臭\end{CJK*}(fetid breath)}~\colorbox{black!3.0688}{\begin{CJK*}{UTF8}{gbsn}要\end{CJK*}(to)}~\colorbox{black!8.14}{\begin{CJK*}{UTF8}{gbsn}吃\end{CJK*}(have)}~\colorbox{black!9.969}{\begin{CJK*}{UTF8}{gbsn}什么\end{CJK*}(which)}~\colorbox{black!13.5}{\begin{CJK*}{UTF8}{gbsn}药\end{CJK*}(medicine)}\\\\
$\circ$~\colorbox{black!3.02}{\begin{CJK*}{UTF8}{gbsn}两\end{CJK*}(two)}~\colorbox{black!4.94}{\begin{CJK*}{UTF8}{gbsn}膝盖\end{CJK*}(knees)}~\colorbox{black!1.94}{\begin{CJK*}{UTF8}{gbsn}里\end{CJK*}(inside)}~\colorbox{black!9.58}{\begin{CJK*}{UTF8}{gbsn}没劲\end{CJK*}( feel weak)}~\colorbox{black!9.44}{\begin{CJK*}{UTF8}{gbsn}怎么回事\end{CJK*}(why)}~\\\\
% \bottomrule
\end{tabular}
\end{adjustbox}
\caption{Confidence scores assigned by the concept encoder on sample queries. A darker color indicates a higher score.}\label{fig::cases}
\end{figure}

%% file: subfiles/5_related_works.tex
\section{Related Works}
\subsection{Medical Query Analysis}
As a growing number of people are posting medical related questions or searching with medical text queries online, researchers have been focusing on new problems and applications based on medical queries or search queries that users generated. \cite{limsopatham2013inferring} analyzes the conceptual relationship in medical records for a better medical search. \cite{stanton2014circumlocution} studies the circumlocution problem in diagnostic medical queries, where users are not able to express their ideas effectively. \cite{zhang2016mining} tries to model user intentions as a classification task for medical text queries. 
% \cite{spink2004study} proposes a technique to improve the spelling suggestion rank in medical text queries. 
\cite{liu2015context} proposes a technique to detect whether users express patient experiences in their medical text queries. 
% \cite{yom2013postmarket} introduces the task of adverse drug reaction discovery from search queries.
In \cite{liu2016augmented}, authors introduce a neural network model to understand users healthcare related questions and try to generate answers appropriately. Being able to infer medical concept transitions from noisy, user-generated healthcare questions may further facilitate various medical applications such as healthcare question-answering, medical dialog systems or recommendation. For example, once we extracted the concept transition $Symptom \to Medicine$ from a question \textit{Any medication is recommended to help me fall asleep easier?}, we may follow up by recommending the user to the nearest pharmacy for further medical consultations on corresponding OTC medicines on Insomnia.

\subsection{Text Classification}
Recently, lots of neural network models are developed for classifying natural language texts into different categories \cite{Xu2014,zhang2015character,Lai2015,Grefenstette2014,zhai2016deepintent}. Those methods achieve decent performance on general text classification tasks.  
The proposed concept transition problem can be cast as a multi-class multi-label classification problem. Unlike traditional text classification tasks like news classification where the existence of some topic words may easily dominate the label for a news title, users tend to mention multiple medical concepts in a single medical text query. It is crucial to extract user medical concept transitions among multiple medical concepts, besides just concept mentions individually.

Also, the aforementioned methods consider the textual information only. With a graph-based formation in this paper, our model is able to seamlessly incorporates an existing concept graph with the medical text query. Moreover, we propose to predict concept mentions as nodes and transitions as links on an abstract level collectively, while most existing works have been focusing on predicting links among concrete entities, e.g. among users in social networks \cite{liben2007link}, or predicting links among entities on a knowledge graph \cite{nickel2016review,bordes2013translating}.

% can be considered as a subgraph link-node collective prediction problem on an abstract level
 
% A recurrent neural network with attention network have been proposed in 
% \cite{zhai2016deepintent} to learn the relationship between a natural language query and bid words in sponsored ads. 

% \subsection{Graph-based link prediction}
% From a broader view, there is another relevant research area that studies link between entities given an information network. 
% Moreover their works considers the textual information only, lea
% From a broader view, there is another relevant research area that studies the quality evaluation
% of question-answer pairs in crowdsourced ques

% For the concept transition inference problem with a graph-based formation, our work can be considered as 

%% file: subfiles/conclusion.tex
\section{Conclusions}
People nowadays are posting or searching with medical text queries extensively on the world wide web. Various medical information needs are expressed diversely in users medical text queries.
% Being able to figure out concept transitions may help us better understand users information search intent and further benefit other tasks such as query rewriting.
In this work, we bring semantic structures to user intention detection in real-world online medical queries by mapping diversely expressed medical queries to a concept graph where each node on a concept graph represents a concept mention and concept transitions are represented as directed edges. 
% We proposed a concept transition inference problem where that detects concept mentions as well as concept transitions collectively. 
A novel neural network structure based on multi-task learning is introduced to extract concept mentions as well as medical concept transitions that users encoded in online healthcare questions collectively. Evaluation results on real-world medical questions address the effectiveness of the proposed model.